\documentclass[tablecaption=bottom,wcp]{jmlr} 

\usepackage{siunitx}

\usepackage{mathtools}
\usepackage{xspace}
\usepackage{amsmath}
\usepackage{bbm}
\usepackage{amssymb}
\usepackage{pifont}
\usepackage{caption}
\usepackage{graphicx}
\usepackage{wrapfig}
\usepackage{enumitem}
\usepackage{multirow}
\usepackage{comment}
\usepackage{tikz}
\usepackage{booktabs} 
\usepackage{cancel}

\jmlrproceedings{}{Preprint under review}

\title{Balancing Simulation-based Inference \\ for Conservative Posteriors}

 \author{
 \Name{Arnaud Delaunoy\nametag{\thanks{Equal contribution}}} \Email{a.delaunoy@uliege.be}\\
 \addr University of Li{\`e}ge
 \AND
 \Name{Benjamin Kurt Miller\nametag{\footnotemark[1]}} \Email{b.k.miller@uva.nl}\\
 \addr University of Amsterdam
 \AND
 \Name{Patrick Forré} \Email{p.d.forre@uva.nl}\\
 \addr University of Amsterdam
 \AND
 \Name{Christoph Weniger} \Email{c.weniger@uva.nl}\\
 \addr University of Amsterdam
 \AND
 \Name{Gilles Louppe} \Email{g.louppe@uliege.be}\\
 \addr University of Li{\`e}ge
 }

\newcommand{\phat}{\hat{p}}
\newcommand{\qhat}{\hat{q}}
\newcommand{\rhat}{\hat{r}}

\newcommand{\alphahat}{\hat{\alpha}}

\newcommand{\varpitilde}{\Tilde{\varpi}}
\newcommand{\pitilde}{\Tilde{\pi}}

\newcommand{\btheta}{\boldsymbol{\theta}}
\newcommand{\bTheta}{\boldsymbol{\Theta}}

\newcommand{\bw}{\boldsymbol{w}}
\newcommand{\bx}{\boldsymbol{x}}

\newcommand{\indicator}{\mathbbm{1}}

\newcommand{\DKL}{\text{KL}}
\newcommand{\Dchi}{\chi^2}
\newcommand{\Mid}{\, \Vert \,}
\renewcommand{\mid}{\, | \,}

\newcommand{\E}{\mathbb{E}}

\DeclareMathOperator*{\argmin}{arg\,min}

\everypar{\looseness=-1}

\begin{document}

\maketitle

\begin{abstract}
Conservative inference is a major concern in simulation-based inference. It has been shown that commonly used algorithms can produce overconfident posterior approximations. Balancing has empirically proven to be an effective way to mitigate this issue. However, its application remains limited to neural ratio estimation. In this work, we extend balancing to any algorithm that provides a posterior density. In particular, we introduce a balanced version of both neural posterior estimation and contrastive neural ratio estimation. We show empirically that the balanced versions tend to produce conservative posterior approximations on a wide variety of benchmarks. In addition, we provide an alternative interpretation of the balancing condition in terms of the $\chi^2$ divergence.
\end{abstract}  


\section{Introduction}
\label{sec:intro}

Simulation-based inference (SBI) \citep{Cranmer2020} is a statistical inference framework that solves the inverse problem of identifying which parameter $\btheta$ generated observation $\bx$ by approximating the posterior $p(\btheta \mid \bx)$ with a surrogate model $\phat(\btheta \mid \bx)$. $\phat$ is constructed from simulated pairs $(\btheta, \bx)$ produced by a generative model where the likelihood $p(\bx \mid \btheta)$ is only implicitly defined. A classic method to produce samples from the surrogate is a rejection sampling technique called Approximate Bayesian Computation \citep{sisson2018handbook}. Recently, there has been significant development of algorithms using machine learning for estimating the posterior \citep{papamakarios2016fast, lueckmann2017flexible, greenberg2019automatic, glockler2021variational, sharrock2022sequential, geffner2022score}, which we call Neural Posterior Estimation (NPE); the likelihood \citep{papamakarios2019sequential, gratton2017glass}; or the likelihood-to-evidence ratio \citep{thomas2016likelihood, tran2017hierarchical, Hermans2019, Durkan2020, miller2021truncated, miller2022contrastive}, which we call Neural Ratio Estimation (NRE).

It has been demonstrated that the estimated surrogate $\phat(\btheta \mid \bx)$ can be more confident than $p(\btheta \mid \bx)$ using common SBI algorithms \citep{hermans2022trust}. This poses a problem for the reliability of SBI in a scientific setting where surrogates must be \emph{conservative}, i.e. avoid inaccurately excluding parameters at a given credibility level. There has been development in testing for overconfidence using empirical expected coverage and related methods \citep{cook2006validation, talts2018validating, hermans2022trust}. \cite{lemos2023sampling} extend expected coverage testing to be a sufficient condition for posterior surrogate correctness, using only samples from the posterior surrogate. In an algorithmic approach to encourage conservativeness, \citet{delaunoytowards} found that the so-called \emph{balance condition} regularizes overconfidence in expectation in surrogates trained with NRE \citep{Hermans2019}. \cite{zhao2021diagnostics} rather test for valid local coverage. \cite{linhart2022validation} focuses on normalizing flows and extends this method to the multivariate setting. In a similar fashion \citep{dalmasso2020confidence, dalmasso2021likelihood, masserano2022simulation} aim to produce valid frequentist coverage.

\paragraph{Contribution}
After providing some background, we generalize the balancing condition to NPE methods and Contrastive Neural Ratio Estimation (NRE-C) \citep{miller2022contrastive}. We provide empirical evidence of the regularizing effect of the balance condition in all of these settings and the first expected coverage tests of NRE-C. Additionally, we relate the balance condition to the $\chi^2$-divergence and generalize it to NPE methods and NRE-C. Code is available at \href{https://github.com/ADelau/balancing_sbi}{\texttt{https://github.com/ADelau/balancing\_sbi}}.

\section{Background}
\label{sec:background}
\paragraph{Posterior estimation (NPE)}
A density estimator $q_{\bw}(\btheta \mid \bx)$ with weights $\bw$, such as a mixture density network \citep{bishop1994mixture} or normalizing flow \citep{papamakarios2019normalizing}, approximates $p(\btheta \mid \bx)$ when the expected Kullback-Leibler divergence
\begin{align}
    \label{eqn:npe-loss}
    \E_{p(\bx)}\left[ \DKL(p(\btheta \mid \bx) \mid q_{\bw}(\btheta \mid \bx) \right],
\end{align}
%
is minimized. In NPE, the surrogate model is directly $\phat(\btheta \mid \bx) \coloneqq q_{\bw}(\btheta \mid \bx)$. 

\paragraph{Ratio estimation (NRE)}
The likelihood-to-evidence ratio $r(\btheta, \bx) \coloneqq \frac{p(\bx \mid \btheta)}{p(\bx)} = \frac{p(\btheta, \bx)}{p(\btheta) p(\bx)}$ is estimated through a supervised learning task using classifier $\varpi(y=1 \mid \btheta, \bx)$. The target conditional distribution $\pi(y=1 \mid \btheta, \bx)$ comes from  $\pi(\btheta, \bx, y) \coloneqq \pi(\btheta, \bx \mid y) \pi(y)$ where the marginals are set to ${\pi(y=0)} \coloneqq {\pi(y=1)} \coloneqq \frac{1}{2}$ and the remaining conditional is defined as
\begin{align}
    \pi(\btheta, \bx \mid y) \coloneqq
    \begin{cases}
        p(\btheta) p(\bx) & y=0 \\
        p(\btheta, \bx) & y=1
    \end{cases}.
\end{align}
The classifier $\varpi(y=1 \mid \btheta, \bx) \coloneqq \sigma \circ f_{\bw}(\btheta, \bx)$ is parameterized by a neural network $f_{\bw}(\btheta, \bx)$ and $\sigma$ is the sigmoid. $\varpi(y=1 \mid \btheta, \bx)$ approximates $\pi(y=1 \mid \btheta, \bx)$ when the NRE loss
\begin{align}
\label{eqn:nre-loss}
    \E_{\pi(\btheta, \bx)} \bigg[ \DKL(\pi(y \mid \btheta, \bx) \Mid \varpi (y \mid \btheta, \bx)) \bigg],
\end{align}
is minimized. Let $\phat(\btheta \mid \bx) \coloneqq \frac{\exp \circ f_{\bw}(\btheta, \bx)}{Z_{\bw}(\bx)} p(\btheta)$, with $Z_{\bw}(\bx) \coloneqq \int \exp \circ  f_{\bw}(\btheta, \bx) p(\btheta) d\btheta$, define the surrogate model. When the loss is zero $f_{\bw}(\btheta, \bx) = \log r(\btheta, \bx)$, recovering the posterior.

Contrastive Neural Ratio Estimation \citep{miller2022contrastive} introduces NRE-C, an algorithm featuring a flexible, multiclass distribution $\pitilde(y)$ for $y = 0, 1, \ldots, K$. 
When the objective \eqref{eqn:nre-c-loss} becomes zero, the NRE-C surrogate model recovers the posterior. 
Details about the method are in \appendixref{appendix:cnre}. NRE-C represents a strict generalization of classifier-based, likelihood-to-evidence ratio estimation methods \citep{Hermans2019, Durkan2020}. 

\paragraph{Conservative surrogates}
Undesirably, simulation-based inference algorithms can produce overconfident surrogate models \citep{hermans2022trust}. We define overconfidence in terms of the $(1-\alpha)$ \emph{expected coverage probability} of the posterior surrogate $\phat(\btheta \mid \bx)$,
\begin{align}
    \label{eqn:expected-coverage-probability}
    1 - \alphahat[\phat; \alpha] \coloneqq \E_{p(\btheta, \bx)} \left[ \indicator(\btheta \in \Theta_{\phat(\btheta \mid \bx)}(1 - \alpha) ) \right],
\end{align}
where $\indicator$ is an indicator function, and $\Theta_{\phat(\btheta \mid \bx)}(1 - \alpha)$ yields the $(1 - \alpha)$ highest posterior density region (HPDR) of $\phat(\btheta \mid \bx)$ with $\alpha \in [0, 1]$. The quantity $(1 - \alpha)$ is called the \emph{nominal coverage probability}. When $\exists \alpha': 1 - \alphahat[\phat; \alpha'] < 1 - \alpha'$, we say that $\phat(\btheta \mid \bx)$ is \emph{overconfident}. Overconfidence is problematic because the surrogate tends to exclude parameter values that are actually plausible at the considered credibility level. On the other hand, extremely \emph{underconfident} surrogates are not informative.
Although there is a tradeoff, scientific applications take a cautious approach by favoring underconfidence. Therefore, we encourage conservative surrogates at credibility level $\alpha'$, which have $1 - \alphahat[\phat; \alpha'] \geq 1 - \alpha'$. One surrogate is more conservative than another when there are more credibility levels at which it is conservative. We show the prior $p(\btheta)$ is conservative in \appendixref{appendix:prior-is-conservative}.

\paragraph{Balance condition}
In an effort to produce conservative surrogates, \citet{delaunoytowards} introduced the \emph{balance condition}. It holds for any classifier $\varpi(y=1 \mid \btheta, \bx)$ that satisfies
\begin{align}
\label{eqn:balance-condition}
\begin{aligned}
    1 = \E_{p(\btheta)p(\bx)}\left[ \varpi(y=1 \mid \btheta, \bx) \right] + \E_{p(\btheta, \bx)} \left[ \varpi(y=1 \mid \btheta, \bx) \right].
\end{aligned}
\end{align}
\citet{delaunoytowards} show that $\E_{p(\btheta, \bx)} \left[ \frac{\pi(y=1 \mid \btheta, \bx)}{\varpi(y=1 \mid \btheta, \bx)} \right] \geq 1$ and $\E_{p(\btheta)p(\bx)} \left[ \frac{\pi(y=0 \mid \btheta, \bx)}{\varpi(y=0 \mid \btheta, \bx)} \right] \geq 1$ for \emph{balanced} classifiers. In expectation, it implies the balanced classifier's probabilities $\varpi(y \mid \btheta, \bx)$ are closer to uniform than the target, encouraging $\rhat(\btheta, \bx)$ to be closer to 1. This brings to surrogate closer to the prior, $p(\btheta)$, which is conservative. 

\section{Extending the balance condition}
\label{sec:extending-balance-condition}

We clarify and generalize the balance criterion: Identifying it with the $\chi^2$-divergence, and applying it to models that can evaluate the (unnormalized) approximate posterior density.

\paragraph{Balance as divergence}
We encourage balance during training by regularizing the loss with a Lagrange multiplier; penalizing solutions that do not satisfy the balance criterion
\begin{align}
    \label{eqn:balance-criterion}
    B[\varpi] \coloneqq B(\bw) \coloneqq
    \left( \E_{p(\btheta)p(\bx)}\left[ \varpi(y=1 \mid \btheta, \bx) \right] + \E_{p(\btheta, \bx)} \left[ \varpi(y=1 \mid \btheta, \bx) \right] - 1 \right)^2,
\end{align}
where $\bw$ are the classifier weights. The effects of balance are clearer when \eqref{eqn:balance-criterion} is rewritten in the form of a $\chi^2$-divergence. The $\chi^2$-divergence \citep{sason2016f} is defined as
\begin{align}
    \label{eqn:chi2-divergence}
    \Dchi(\pi(y) \Mid \varpi(y)) \coloneqq \int \left( \frac{\varpi(y)}{\pi(y)} - 1 \right)^2 \pi(y) \, dy,
\end{align}
where $\varpi(y) \coloneqq \int \varpi(y \mid \btheta, \bx) \pi(\btheta, \bx) d\btheta \, d\bx$ is the marginal classifier. With intermediate steps shown in \eqref{eqn:balance-is-chi2}, we identify that $B[\varpi] = \Dchi(\pi(y) \Mid \varpi(y))$! Therefore, a balanced classifier satisfies $\varpi(y) = \pi(y)$, i.e. balancing regularizes the marginal classifier towards the target distribution for $y$. We explore the balance condition for a multiclass $y$ in \appendixref{appendix:balance-in-terms-of-f-div}.

\paragraph{Balance criterion for alternative models}
The balance criterion regularizes marginal distribution $\varpi(y)$; this makes sense for NRE which defines $\varpi(y)$ and target marginal $\pi(y)$.
However, we are interested in regularizing objectives $L(\bw)$ that either do not introduce a binary auxiliary variable $y$, or use an alternative. In order to apply the balance criterion, we propose to use the same target distribution $\pi(\btheta, \bx, y)$ as NRE does and define a classifier in terms of the variational (unnormalized) posterior approximant $\qhat_{\bw}(\btheta \mid \bx)$. We approximate $r(\btheta, \bx) \coloneqq \frac{p(\btheta, \bx)}{p(\btheta)p(\bx)} = \frac{p(\btheta \mid \bx)}{p(\btheta)}$ with $\frac{\qhat_{\bw}(\btheta \mid \bx)}{p(\btheta)}$ which yields the classifier $\varpi(y=1 \mid \btheta, \bx; \qhat_{\bw}) \coloneqq \frac{\qhat_{\bw}(\btheta \mid \bx) / p(\btheta)}{1+\qhat_{\bw}(\btheta \mid \bx) / p(\btheta)}$ and regularize $L(\bw)$ with $B(\bw)$ and Lagrange multiplier $\lambda$
\begin{equation}
    \label{eqn:arbitrary-balanced-loss}
    L(\bw) + \lambda B(\bw).
\end{equation}
The main contribution is reformulating $B(\bw)$ to be expressed in terms of $\qhat_{\bw}$, which generalizes the balance condition to models which allow for approximate density evaluation!
We consider losses $L(\bw)$ that go to zero if and only if $\qhat_{\bw}(\btheta \mid \bx) = p(\btheta \mid \bx)$. When this is true, the balance condition \eqref{eqn:balance-condition} also holds since $B(\bw)$ becomes zero:
\begin{align}
\begin{aligned}
    B(\bw) &\coloneqq \left( \int (\pi(\btheta, \bx \mid y=0) + \pi(\btheta, \bx \mid y=1)) \varpi(y=1 \mid \btheta, \bx; \qhat_{\bw}) d\btheta \, d\bx - 1 \right)^2 \\
    &= \left(\int (p(\btheta)p(\bx) + p(\btheta, \bx)) \frac{\qhat_{\bw}(\btheta \mid \bx)/p(\btheta)}{1 + \qhat_{\bw}(\btheta \mid \bx)/p(\btheta)} d\btheta \, d\bx - 1 \right)^2 \\
\end{aligned}
\end{align}
That means for loss $L(\bw) = 0$ the regularized version \eqref{eqn:arbitrary-balanced-loss} also goes to zero; just like BNRE.

\paragraph{Balanced Neural Posterior Estimation (BNPE)}
We propose BNPE which regularizes NPE's objective \eqref{eqn:npe-loss} with the balance criterion to train a normalized density estimator. We have $\qhat_{\bw}(\btheta \mid \bx) \coloneqq q_{\bw}(\btheta \mid \bx)$. The corresponding classifier becomes $\varpi(y=1 \mid \btheta, \bx; \qhat_{\bw}) \coloneqq \frac{q_{\bw}(\btheta \mid \bx) / p(\btheta)}{1+q_{\bw}(\btheta \mid \bx) / p(\btheta)}$. Training minimizes \eqref{eqn:arbitrary-balanced-loss}, substituting the appropriate loss and classifier.

\paragraph{Balanced Contrastive Neural Ratio Estimation (BNRE-C)}
We propose BNRE-C which regularizes NRE-C's objective \eqref{eqn:nre-c-loss} with the balance criterion to train a ratio estimator. The density estimator is $\qhat_{\bw}(\btheta \mid \bx) \coloneqq \exp \circ h_{\bw}(\btheta, \bx) p(\btheta)$. The corresponding classifier becomes $\varpi(y=1 \mid \btheta, \bx; \qhat) \coloneqq \frac{\exp \circ h_{\bw}(\btheta , \bx) p(\btheta) / p(\btheta)}{1+\exp \circ h_{\bw}(\btheta , \bx) p(\btheta) / p(\btheta)} = \sigma \circ h_{\bw}(\btheta, \bx)$. Training minimizes \eqref{eqn:arbitrary-balanced-loss}, substituting the appropriate loss and classifier.

\section{Experiments}
\label{sec:experiments}
\begin{figure}[htb]
    \vspace{-1.4em}
    \centering
    \includegraphics[width=\textwidth]{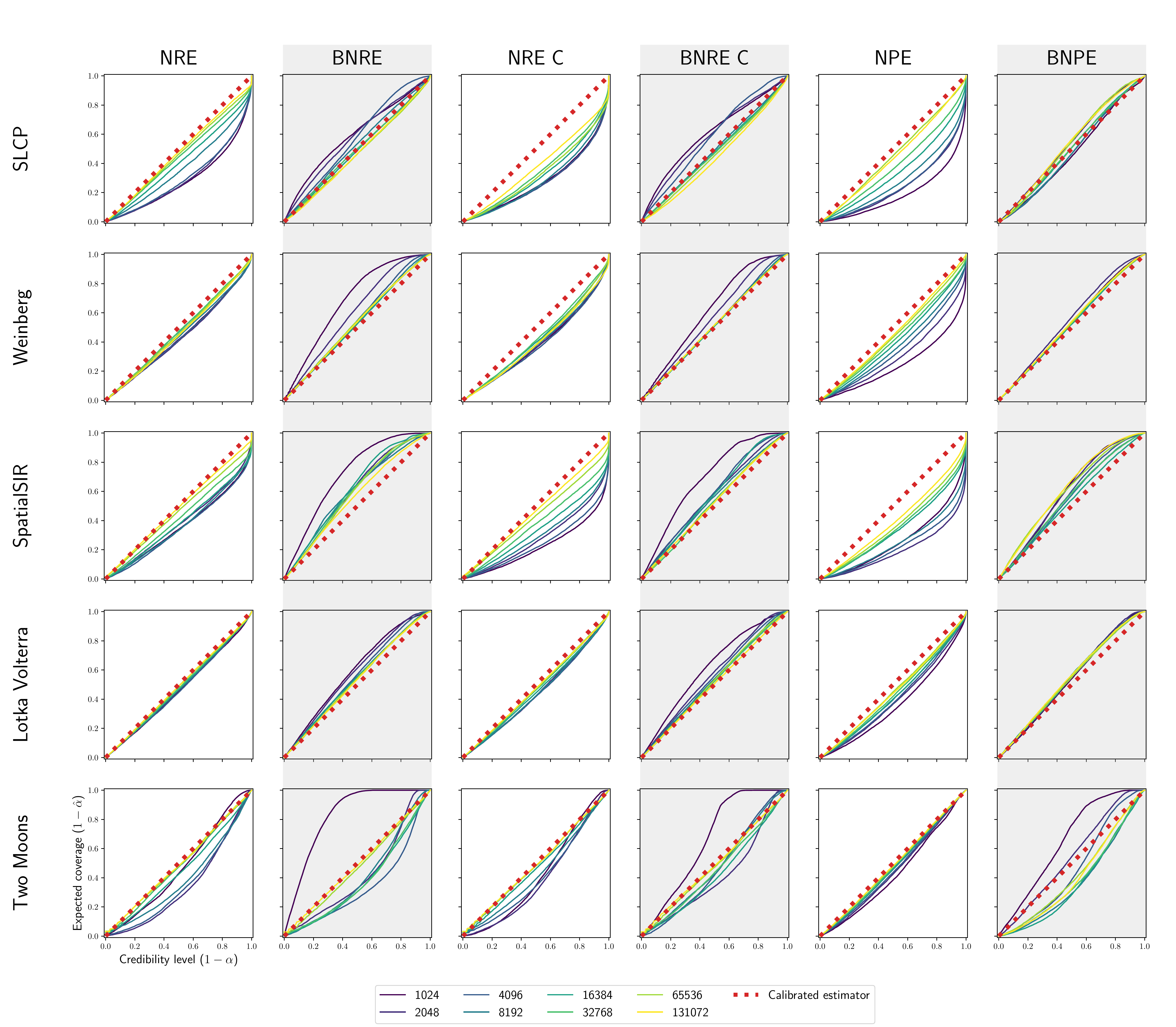}
    \caption{Empirical expected coverage for increasing simulation budgets. A perfectly calibrated surrogate has an expected coverage probability equal to the nominal coverage probability and 
    produces a diagonal line. A conservative surrogate has an expected coverage curve at or above the diagonal line.
    An overconfident surrogate produces curves below the diagonal line.
    Balanced algorithms tend to produce conservative surrogates. 
    5 runs are performed for each simulation budget with the median $\alphahat$ at each nominal credibility reported.
    }
    \label{fig:coverage}
\end{figure}

We investigate whether our proposed algorithms BNPE and BNRE-C lead to more conservative surrogate models compared to their non-balanced counterparts. Our diagnostics test the empirical expected coverage probability \eqref{eqn:expected-coverage-probability} of considered algorithms trained with ($\lambda > 0$) and without ($\lambda = 0$) the balance criterion. We also benchmark BNRE \citep{delaunoytowards}, an algorithm that trains a neural network to minimize \eqref{eqn:arbitrary-balanced-loss} using \eqref{eqn:nre-loss} as $L(\bw)$. The empirical expected coverage tests are performed on a set of benchmarks of varying difficulty. Simulator descriptions can be found in \appendixref{sec:benchmarks} and neural network architectures in \appendixref{sec:architectures}. Expected coverage for these benchmarks are shown in \figureref{fig:coverage}.

For both BNPE and BNRE-C, enforcing the balance criterion tends to produce conservative surrogates on most of the benchmarks, just as it did with BNRE. The only exception was Two Moons, which consistently produced overconfident surrogates for all algorithms. Taken as a whole, this provides some evidence that regularizing models allowing posterior density evaluation with the balance criterion makes them more conservative.
Our results suggest that enforcing the balance criterion did not negatively impact the informativeness of the surrogate for higher simulation budgets.
This is quantified in \appendixref{sec:additional_experiments}  where we estimate the information contained in the surrogates and the empirical balance error. A qualitative analysis of the posteriors is provided in Appendix \ref{sec:qualitative}. 

We observed that normalizing flows struggle to minimize the balance criterion on the SLCP and Weinberg benchmarks for low simulation budgets. Normalizing flows must learn the prior as a multiplicative component in their density estimate, while likelihood-to-evidence ratio methods use the ground truth prior to construct the surrogate. We hypothesize that this difference may play a role in the observed behavior. We discuss this in \appendixref{sec:refining} and empirically reduce balance error by initializing the flow close to the prior.

\section{Conclusions}

In this work, we have shown that balancing can be applied to any simulation-based inference algorithm that yields a posterior density estimator, hence extending the applicability of balancing. We have also shown that the balancing condition can be expressed as a $\chi^2$ divergence. We believe that this reformulation is a stepping stone towards a better understanding of the balancing condition and could inspire novel algorithms.

Let us note that there exist algorithms that do not directly provide the posterior density and hence do not fall into this framework. Some algorithms only aim to provide an unnormalized posterior approximation. This is the case of algorithms that model the likelihood \citep{papamakarios2019sequential} or an unnormalized version of the likelihood-to-evidence ratio \citep{Durkan2020}. A direct consequence is that those algorithms are not necessarily balanced at optimum. Enforcing the balancing condition may hence prevent reaching the optimal solution. Some algorithms also only allow sampling from the posterior surrogate. This is the case of score-based methods \citep{song2020score, geffner2022score}. In such a setting, the balancing condition cannot be computed. Future work would then include a reformulation of the balancing condition that uses only samples from the posterior surrogate.

It should also be noted that enforcing that the marginal classification matches the prior over classes is something that can be extended to more than two classes. The $\chi^2$ formulation of the balancing condition could hence be extended to multi-class classification methods such as \cite{Durkan2020} and \cite{miller2022contrastive} as shown in Appendix \ref{appendix:balance-in-terms-of-f-div}. Whether the effect of this regularization on the posterior would be the same as in the two-classes setting remains to be studied and is left for future work.

\acks{
Arnaud Delaunoy would like to thank the National Fund for Scientific Research (F.R.S.-FNRS) for his scholarship. 
Benjamin Kurt Miller and Gilles Louppe collaborate together as part of the ELLIS PhD program, receiving travel support from the ELISE mobility program which has received funding from the European Union’s Horizon 2020 research and innovation programme under ELISE grant agreement No 951847.
Christoph Weniger received funding from the European Research Council (ERC) under the European Union’s Horizon 2020 research and innovation programme (Grant agreement No. 864035 -- UnDark).
}

\clearpage

\bibliography{bibliography}

\clearpage

\appendix

\section{The Prior is Conservative}
\label{appendix:prior-is-conservative}

In this section, we prove that the prior $p(\btheta)$ is conservative at credibility level $\alpha$. We begin by gathering the necessary facts and definitions. First, we give a definition of the function 
\begin{align}
    \label{eqn:get-hpdr-func}
    &\Theta_{\phat(\btheta \mid \bx)}(1 - \alpha) \coloneqq \bar{\Theta}_{\alpha} \in \argmin_{\bar{\Theta}} \int_{\bar{\Theta}} d\btheta
    &\text{s.t.}&
    &\int_{\bar{\Theta}} \phat(\btheta \mid \bx) d\btheta = 1 - \alpha
\end{align}
which yields the $(1 - \alpha)$ HPDR of $\phat(\btheta \mid \bx)$, denoted $\bar{\Theta}_{\alpha}$, with $\alpha \in [0, 1]$ and $\bar{\Theta}$ denoting a measure space. Recall, the $(1-\alpha)$ expected coverage probability of surrogate $\phat(\btheta \mid \bx)$:
\begin{align*}
    1 - \alphahat[\phat; \alpha] \coloneqq \E_{p(\btheta, \bx)} \left[ \indicator(\btheta \in \Theta_{\phat(\btheta \mid \bx)}(1 - \alpha) )\right] = \int_{\bar{\Theta} \times \bar{\mathcal{X}}} p(\btheta, \bx) \indicator(\btheta \in \Theta_{\phat(\btheta \mid \bx)}(1 - \alpha)) d\btheta \, d\bx,
\end{align*}
which was defined in \eqref{eqn:expected-coverage-probability}, and where $\bar{\mathcal{X}}$ denotes another measure space.
Finally, we say that posterior surrogate $\phat(\btheta \mid \bx)$ is conservative at credibility level $\alpha$ when
\begin{align}
    1 - \alphahat[\phat; \alpha] \geq 1 - \alpha.
\end{align}

For the proof we let $\phat(\btheta \mid \bx) \coloneqq p(\btheta)$. The $(1-\alpha)$ expected coverage probability of $p(\btheta)$ is
\begin{align}
\begin{aligned}
    \label{eqn:prior-expected-coverage}
    1 - \alphahat[p(\btheta); \alpha] 
    &= \int_{\bar{\Theta} \times \bar{\mathcal{X}}} p(\btheta, \bx) \indicator(\btheta \in \Theta_{p(\btheta)}(1 - \alpha)) d\btheta \, d\bx \\
    &= \int_{\bar{\Theta}} p(\btheta) \indicator(\btheta \in \Theta_{p(\btheta)}(1 - \alpha)) d\btheta \\
    &= \int_{\bar{\Theta}_{\alpha}} p(\btheta) d\btheta,
\end{aligned}
\end{align}
where we first integrated over $\bx$ then applied the indicator function to the bounds of integration, reducing to $\bar{\Theta}_{\alpha}$ by \eqref{eqn:get-hpdr-func}. We substitute the region of integration into the constraint for our surrogate $p(\btheta)$ and recover
\begin{align}
    \int_{\bar{\Theta}_{\alpha}} \phat(\btheta \mid \bx) d\btheta = \int_{\bar{\Theta}_{\alpha}} p(\btheta) d\btheta = 1 - \alpha.
\end{align}
Our constraint is equal to the last line of \eqref{eqn:prior-expected-coverage}, implying
\begin{align}
\begin{aligned}
    1 - \alphahat[p(\btheta); \alpha] = 1 - \alpha \geq 1 - \alpha,
\end{aligned}
\end{align}
which means that $p(\btheta)$ has exact coverage at every credibility level $\alpha$, which is conservative.

\clearpage

\section{Contrastive Neural Ratio Estimation}
\label{appendix:cnre}
Here we introduce a summary of NRE-C \citep{miller2022contrastive} in our notation and indicate the specific hyperparameter choice in the experiments. First, we define the target distribution for the supervised learning problem. We let marginal distribution $\pitilde(y)$ with $y \in \{0, 1, \ldots, K\}$ have probabilities $\pitilde(y=k) \coloneqq \pitilde_{K}$ for all $k \geq 1$ and $\pitilde(y=0) \coloneqq \pitilde_{0}$ which yields the relationship $\pitilde_0 = 1 - K \pitilde_{K}$. The remaining conditional is
\begin{align}
    \pitilde(\bTheta, \bx \mid y = k) &\coloneqq
    \begin{cases}
        p(\btheta_1) \cdots p(\btheta_K) p(\bx) & k=0 \\
    	p(\btheta_1) \cdots p(\btheta_K) p(\bx \mid \btheta_k) & k = 1, \ldots, K
    \end{cases},
\end{align}
where $\bTheta \coloneqq (\btheta_1, ..., \btheta_K)$ are contrastive parameters, sampled from the prior $p(\btheta)$. We fit the variational, multiclass classifier 
\begin{align}
    \varpitilde(y = k \mid \bTheta, \bx) &\coloneqq
    \begin{cases}
    	\frac{K}{K + \sum_{i=1}^{K} \exp \circ h_{\bw}(\btheta_i, \bx)} & k = 0 \\
    	\frac{\exp \circ h_{\bw}(\btheta_k,\bx))}{K + \sum_{i=1}^{K} \exp \circ h_{\bw}(\btheta_i,\bx)} & k = 1, \ldots, K
    \end{cases}
\end{align}
by minimizing $\E_{\pitilde(\bTheta, \bx)} \left[ \DKL(\pitilde(y \mid \bTheta, \bx) \mid \varpitilde(y \mid \bTheta, \bx)) \right]$, where $h_{\bw}$ is a neural network parameterized by weights $\bw$. We write the loss function out explicitly:
\begin{align}
    \label{eqn:nre-c-loss}
    \begin{split}
    L[\varpitilde] \coloneqq &-\frac{1}{1 + \gamma} \mathbb{E}_{\pitilde(\bTheta, \bx \mid y=0)} \left[
    \log \varpitilde(y =0 \mid \bTheta, \bx) 
    \right] \\
    &- \frac{\gamma}{1 + \gamma} \mathbb{E}_{\pitilde(\bTheta, \bx \mid y=K)} \left[ 
    \log \varpitilde(y = K \mid \bTheta, \bx) 
    \right],
    \end{split}
\end{align}
where we introduced $\gamma \coloneqq \frac{\pitilde(y \geq 1)}{\pitilde(y=0)} = \frac{K \pitilde_{K}}{\pitilde_{0}}$. There are only two terms in this sum because we exploited symmetries in the conditionals $\pitilde(\bTheta, \bx \mid y = k)$ when $k \neq 0$. We define the surrogate model as $\phat(\btheta \mid \bx) \coloneqq \frac{\exp \circ h_{\bw}(\btheta,\bx)}{Z_{\bw}(\bx)} p(\btheta)$ with $Z_{\bw}(\bx) \coloneqq \int \exp \circ h_{\bw}(\btheta,\bx) p(\btheta) \, d\btheta$.

In the experiments we took $\gamma \coloneqq 1$ and $K \coloneqq 5$.

In \sectionref{sec:extending-balance-condition}, we introduce the balance criterion to NRE-C's loss function. We emphasize here that the $B(\bw) \coloneqq B[\varpi]$ term is a functional of the \emph{binary classifier} $\varpi \coloneqq \sigma \circ h_{\bw}(\btheta, \bx)$ parameterized by the same neural network as the multiclass $\varpitilde$.

\clearpage

\section{Balance condition in terms of f-divergences}
\label{appendix:balance-in-terms-of-f-div}

\paragraph{Balance and $\chi^2$-divergence}
$\Dchi$ is an f-divergence, as defined in \eqref{eqn:chi2-divergence}. The steps to show the connection between $\Dchi$ and the balancing condition are as follows:
\begin{align}
\label{eqn:balance-is-chi2}
\begin{aligned}
    \Dchi(\pi(y) \Mid \varpi(y)) &\coloneqq \int \left( \frac{\varpi(y)}{\pi(y)} - 1 \right)^2 \pi(y) \, dy \\
    &= \left( \frac{\varpi(y=0)}{\pi(y=0)} - 1 \right)^2 \pi(y=0) + \left(\frac{\varpi(y=1)}{\pi(y=1)} - 1 \right)^2 \pi(y=1) \\
    &= \frac{(\varpi(y=0)-\pi(y=0))^2}{\pi(y=0)} + \frac{(\varpi(y=1) - \pi(y=1))^2}{\pi(y=1)} \\
    &= \frac{(\varpi(y=1)-\pi(y=1))^2}{\pi(y=0)} + \frac{(\varpi(y=1) - \pi(y=1))^2}{\pi(y=1)} \\
    &= (\varpi(y=1)-\pi(y=1))^2 \left( \frac{1}{\pi(y=0)} + \frac{1}{\pi(y=1)} \right) \\
    &= 4 (\varpi(y=1)-\pi(y=1))^2 \\
    &= (2 \varpi(y=1) - 1)^2 \\
    &= \left( 2 \int \varpi(y=1 \mid \btheta, \bx) \pi(\btheta. \bx) d\btheta \, d\bx - 1 \right)^2 \\
    &= \left( 2 \int \varpi(y=1 \mid \btheta, \bx) \left( \int \pi(\btheta, \bx \mid y) \pi(y) dy \right) d\btheta \, d\bx - 1 \right)^2 \\
    &= \left( \int \varpi(y=1 \mid \btheta, \bx) (\pi(\btheta, \bx \mid y=0) + \pi(\btheta, \bx \mid y=1)) d\btheta \, d\bx - 1 \right)^2 \\
    &= \left( \E_{p(\btheta)p(\bx)}\left[ \varpi(y=1 \mid \btheta, \bx) \right] + \E_{p(\btheta, \bx)} \left[ \varpi(y=1 \mid \btheta, \bx) \right] - 1 \right)^2
\end{aligned}
\end{align}
where we used $\pi(y=0)=\pi(y=1)=\frac{1}{2}$ and 
$\varpi(\btheta, \bx) \coloneqq \pi(\btheta, \bx)$.

\paragraph{Kullback–Leibler divergence chain rule}
An important result from \citet{delaunoytowards} was that the optimal classifier is balanced. It can be seen from the chain rule:
\begin{align}
\begin{aligned}
    \label{eqn:kld-chain-rule}
    \DKL(\pi(\btheta, \bx, y) \Mid \varpi(\btheta, \bx, y)) &= 
    \underbrace{\DKL\left( \pi(\btheta, \bx) \Mid \varpi(\btheta, \bx) \right)}_{= 0} + 
    \E_{\pi(\btheta, \bx)} \bigg[ \DKL(\pi(y \mid \btheta, \bx) \Mid \varpi (y \mid \btheta, \bx) \bigg] \\
    &= \DKL\left( \pi(y) \Mid \varpi(y) \right) + 
    \E_{\pi(y)} \bigg[ \DKL(\pi(\btheta, \bx \mid y) \Mid \varpi (\btheta, \bx \mid y) \bigg],
\end{aligned}
\raisetag{1.2\baselineskip}
\end{align}
since $\varpi(\btheta, \bx) \coloneqq \pi(\btheta, \bx)$. Minimizing the left hand side of this equation also minimizes the right hand side with a one-to-one trade off between terms. When $\varpi(y \mid \btheta, \bx) = \pi(y \mid \btheta, \bx)$, the left left hand side of the equation becomes zero, implying that every term on the right hand side becomes zero as well. The balance condition enforced $\varpi(y) = \pi(y)$, which holds when the left hand side is zero, due to $\DKL\left( \pi(y) \Mid \varpi(y) \right) = 0$; therefore, $\pi(y \mid \btheta, \bx)$ is balanced and enforcing the balance condition does \emph{not} exclude this optimal classifier, i.e. the target distribution, from our solution set!

Consider the effects of including the balance condition in the loss function. Given Lagrange multiplier $\lambda$, the loss becomes
\begin{align}
    \label{eqn:balanced-loss}
    L[\varpi] + \lambda B[\varpi] = \E_{\pi(\btheta, \bx)} \bigg[ \DKL(\pi(y \mid \btheta, \bx) \Mid \varpi (y \mid \btheta, \bx) \bigg] + \lambda \Dchi\left( \pi(y) \Mid \varpi(y) \right).
\end{align}
The balance criterion \emph{effectively} puts extra weight on the ``balance term'' in the unregularized objective \eqref{eqn:kld-chain-rule}. This statement is not strictly true because the functional form of $\DKL$ and $\Dchi$ are different; however, they both become zero when $\varpi(y) = \pi(y)$ and increase when $\varpi(y)$ becomes increasingly ``different'' from $\pi(y)$. The first claim is based on the inequality $\DKL(\pi\|\varpi) \le \Dchi(\pi\|\varpi)$, which is shown in \eqref{eqn:chi2-kl-inequality}. The second is a property of divergences.

\paragraph{Explicit BNRE loss definition}
Together this yields our regularized sample approximation of the loss for BNRE under combination of the objective and regularizer
\begin{align}
\begin{aligned}
    \ell(\bw) &= \E_{\pi(\btheta, \bx)} \bigg[ \DKL(\pi(y \mid \btheta, \bx) \Mid \varpi (y \mid \btheta, \bx) \bigg] + \lambda \Dchi\left( \pi(y) \Mid \varpi(y) \right) \\
    &\begin{aligned}
    &\approx - \frac{1}{2B} \left[- \sum_{b=1}^{B} \log \left(
        1 - \sigma \circ h_{\bw}(\btheta^{(b)}, \bx^{(b)})
    \right) 
    + \sum_{b'=1}^{B}\log \left(
        \sigma \circ h_{\bw}(\btheta^{(b')}, \bx^{(b')})
    \right) \right] \\
    & \hspace{2em} + \lambda \left[ \sum_{b=1}^{B} \sigma \circ h_{\bw}(\btheta^{(b)}, \bx^{(b)}) + \sum_{b'=1}^{B}\sigma \circ h_{\bw}(\btheta^{(b')}, \bx^{(b')}) - 1 \right]^2,
    \end{aligned}
\end{aligned}
\end{align}
where $\btheta^{(b)}, \bx^{(b)} \sim p(\btheta) p(\bx)$ and $\btheta^{(b')}, \bx^{(b')} \sim p(\btheta, \bx)$.

\paragraph{Balance as a Kullback-Liebler divergence}
Rather than using the balance criterion, it is also possible to apply the Kullback-Liebler divergence to regularize $\varpi(y)$:
\begin{align}
\begin{aligned}
    &\DKL(\pi(y) \Mid \varpi(y)) \coloneqq \int \pi(y) \log \frac{\pi(y)}{\varpi(y)} \, dy \\
    &\quad = \frac{1}{2} \int \left( \log \pi(y=0) - \log \varpi(y=0) + \log \pi(y=1) - \log \varpi(y=1) \right) \, dy \\
    &\quad = - \frac{1}{2} \int \left( \log \varpi(y=0) + \log \varpi(y=1) \right) \, dy - \log 2 \\
    \begin{split}
        &\quad = - \frac{1}{2} \int \Bigg[ \log \left( \int \varpi(y=0 \mid \btheta, \bx) \left( \int \pi(\btheta, \bx \mid y) \pi(y) dy \right) d\btheta \, d\bx  \right) \\
        & \hspace{2em} + \log \left( \int \varpi(y=1 \mid \btheta, \bx) \left( \int \pi(\btheta, \bx \mid y) \pi(y) dy \right) d\btheta \, d\bx \right) \Bigg] \, dy - \log 2 
    \end{split}\\
    \begin{split}
        &\quad = - \frac{1}{2} \int \Bigg[ \log \left( \frac{1}{2} \int \varpi(y=0 \mid \btheta, \bx) \left( \pi(\btheta, \bx \mid y=0) + \pi(\btheta, \bx \mid y=1) \right) d\btheta \, d\bx  \right) \\
        & \hspace{2em} + \log \left( \frac{1}{2} \int \varpi(y=1 \mid \btheta, \bx) \left( \pi(\btheta, \bx \mid y=0) + \pi(\btheta, \bx \mid y=1) \right) d\btheta \, d\bx \right) \Bigg] \, dy \\
        & \hspace{2em} - \log 2.
    \end{split}
\end{aligned}
\end{align}
The objective is motivated information theoretically because it appears in the chain rule for the Kullback-Leibler divergence, like in \eqref{eqn:kld-chain-rule}, but it is unappealing from an optimization perspective: The $\log$ of the integrals leads to biased empirical estimates based-on samples.

\paragraph{Kullback-Leibler and $\chi^2$ divergences}
For probability measures $P$ and $Q$ on measure space $\mathcal{X}$, the $\chi^2$ divergence is
\begin{align}
    \chi^2(P \Mid Q) \coloneqq \int_{\mathcal{X}} \left(\frac{dP}{dQ} - 1\right)^2 dQ
= \int_{\mathcal{X}} \left[ \left(\frac{dP}{dQ}\right)^2 - 1\right]dQ.
\end{align}
Therefore,
\begin{align}
    \DKL(P \Mid  Q) =  \int_{\mathcal{X}} \ln\frac{dP}{dQ} dP
    \leq  \int_{\mathcal{X}} \left(\frac{dP}{dQ} - 1\right) dP = \int_{\mathcal{X}} \left[ \left(\frac{dP}{dQ}\right)^2 - 1 \right]dQ
\end{align}
where we used the concavity of the logarithm $\ln x \leq x-1, \forall x >0$. This proves
\begin{align}
    \label{eqn:chi2-kl-inequality}
    \DKL(P \Mid  Q) \leq \chi^2(P \Mid Q), \quad \forall P\ll Q.
\end{align}
Alternative proofs and other relevant inequalities can be found in the materials \citet{sason2016f, dragomir2002upper, su1995methods, su2002choosing}. The consequence of this is that if we minimize the $\chi^2$-divergence, we will Kullback-Leibler divergence as well.

\clearpage

\paragraph{Balancing multiclass objectives}
Since the balance condition can be specified as a divergence, we can define a multiclass balance conditions for ratio estimators featuring arbitrary multiclass $y$ variables such as those introduced by \citet{Durkan2020} and \citet{miller2022contrastive}. Additionally, the $\chi^2$-divergence formulation enables regularization of classifiers fitting to $\pitilde(y \mid \bTheta, \bx)$ that have arbitrary marginal distributions $\pitilde(y)$. We make no statements about the effects changing $\pitilde(y)$ might have on the estimated posterior $\phat(\btheta \mid \bx)$.
\begin{align}
\label{eqn:multiclass-balance-criterion}
\begin{aligned}
    &\Dchi(\pitilde(y) \Mid \varpitilde(y)) \coloneqq \int \left( \frac{\varpitilde(y)}{\pitilde(y)} - 1 \right)^2 \pitilde(y) \, dy \\
    &= \sum_{i=0}^{K} \left( \frac{\varpitilde(y=i)}{\pitilde(y=i)} - 1 \right)^2 \pitilde(y=i) \\
    &= \sum_{i=0}^{K} \frac{(\varpitilde(y=i) - \pitilde(y=i))^2}{\pitilde(y=i)} \\
   &= (K+1) \; \sum_{i=0}^{K} \left(\varpitilde(y=i) - \frac{1}{K+1}\right)^2 \qquad \text{(uniform $\pitilde(y)$)} \\
    &= \frac{1}{K+1} \sum_{i=0}^{K} \left((K+1) \varpitilde(y=i) - 1\right)^2 \\
    &=  \frac{1}{K+1} \sum_{i=0}^{K} \left( \int (K+1) \varpitilde(y=i \mid \bTheta, \bx) \left( \sum_{j=0}^{K} \pitilde(\bTheta, \bx | y=j) \pitilde(y=j) \right) d\bTheta d\bx - 1 \right)^2 \\
    &= \frac{1}{K+1} \sum_{i=0}^{K} \left( \int \varpitilde(y=i \mid \bTheta, \bx) \left( \sum_{j=0}^{K} \pitilde(\bTheta, \bx \mid y=j) \right) d\bTheta \, d\bx - 1 \right)^2 \quad \text{(uniform $\pitilde(y)$)}
\end{aligned}
\end{align}
where $\varpitilde(\bTheta, \bx, y) \coloneqq \varpitilde(y \mid \bTheta, \bx) \pitilde(\bTheta, \bx)$ and $\varpitilde(y \mid \bTheta, \bx)$ is a variational multiclass classifier. Some problems may offer optimization using symmetries in the definition of the sampling distribution $\pitilde(\bTheta, \bx \mid y=j)$. Here we followed the NRE-C convention that $y = 0, 1, \ldots, K$.

We want to note here that we did not apply \eqref{eqn:multiclass-balance-criterion} in our experiments, rather we applied \eqref{eqn:balance-criterion} even to the applications featuring a multiclass $y$. The analysis about the effectiveness of a multiclass regularizer versus the binary balance criterion are left for future work.

\clearpage

\section{Benchmarks description}
\label{sec:benchmarks}

The \emph{SLCP} simulator models a fictive problem with 5 parameters. The observable $\bx$ is composed of 8 scalars which represent the 2D-coordinates of 4 points. 
The coordinate of each point is sampled from the same multivariate Gaussian whose mean and covariance matrix are parametrized by $\btheta$. We consider an alternative version of the original task \citep{papamakarios2019sequential} by inferring the marginal posterior density of 2 of those parameters. In contrast to its original formulation, the likelihood is not tractable due to the marginalization.

The \emph{Weinberg} problem \citep{weinberg} concerns a simulation of high energy particle collisions $e^+e^- \to \mu^+ \mu^-$. The angular distributions of the particles can be used to measure the Weinberg angle $\bx$
in the standard model of particle physics, leading to an observable composed of 20 scalars. From the scattering angle, we are interested in inferring Fermi's constant $\btheta$.

The \emph{Spatial SIR} model \citep{hermans2022trust} involves a grid-world of susceptible,
infected, and recovered individuals. Based on initial conditions and the infection and recovery rate $\btheta$,
the model describes the spatial evolution of an infection.
The observable $\bx$ is a snapshot of the grid-world after some fixed amount of time. The grid used is of size 50 by 50. 

The \emph{Lotka-Volterra} population model \citep{lotka,volterra1926fluctuations} describes a process of interactions between a predator and a prey species. The model is conditioned on 4 parameters $\btheta$ which influence the reproduction and mortality rate of the predator and prey species. We infer the marginal posterior of the predator parameters from time series of $2001$ steps representing the evolution of both populations over time. The specific implementation is based on a Markov Jump Process as in \citet{papamakarios2019sequential}.

The \emph{Two Moons} \citep{greenberg2019automatic} simulator models a fictive problem with 2 parameters. The observable $\bx$ is composed of 2 scalars which represent the 2D-coordinates of a random point sampled from a crescent-shaped distribution shifted and rotated around the origin depending on the parameters' values. Those transformations involve the absolute value of the sum of the parameters leading to a second crescent in the posterior and hence making it multi-modal.

\section{Architectures and hyper-parameters}
\label{sec:architectures}

Table \ref{tab:architectures} summarizes the architectures and hyper-parameters used for each benchmark. The architectures are separated into two parts: the embedding and the head networks. The embedding network compresses the observable into a set of features. The head network then uses those features concatenated with the parameters to produce the target quantity. The head can either be a classifier or a normalizing flow depending on the algorithm considered. The learning rate is scheduled during training. Table \ref{tab:architectures} provides the initial learning rates. Those are then divided by $10$ each time no improvement was observed on the validation loss for $10$ epochs.

\begin{table}[h]
    \centering
    \begin{tabular}{lllllll}
        \toprule
        & SLCP & Weinberg & Lotka-V. & Spatial SIR & Two Moons \\
        \midrule
        \emph{Embedding network} & None &  None & CNN & CNN & None\\
        \emph{Embedding layers} & / &  / & $8$ & $8$ & / \\ 
        \emph{Embedding channels} & / &  / & $8$ & $16$ & / \\ 
        \emph{Convolution type} & / & / & Conv1D & Conv2D & / \\ 
        \emph{Classifier Head} & MLP & MLP & MLP & MLP & MLP \\
        \emph{Classifier layers} & $6$ & $6$ & $6$ & $6$ & $6$ \\
        \emph{Classifier hidden neurons} & $256$ &  $256$ & $256$ & $256$  & $256$ \\
        \emph{Flow Head} & NSF & NSF & NSF & NSF & NSF \\
        \emph{Flow layers} & $3$ & $3$ & $3$ & $3$ & $3$ \\
        \emph{Flow hidden neurons} & $256$ &  $256$ & $256$ & $256$ & $256$ \\
        \emph{Learning rate} & $0.001$ &  $0.001$ & $0.001$ & $0.001$ & $0.001$ \\ 
        \emph{Epochs} & $500$ &  $500$ & $500$ & $500$ & $500$ \\ 
        \emph{Batch size} & $256$ & $256$ & $256$ & $256$ & $256$ \\ 
        \emph{$\lambda$ (balanced algorithms)} & $100$ & $100$ & $100$ & $100$ & $100$ \\
        \emph{$\gamma$ (NRE-C)} & $1$ & $1$ & $1$ & $1$ & $1$ \\
        \emph{$K$ (NRE-C)} & $5$ & $5$ & $5$ & $5$ & $5$ \\
        \bottomrule
    \end{tabular}
    \caption{Architectures and training hyper-parameters}
    \label{tab:architectures}
\end{table}

\clearpage

\section{Additional experiments}
\label{sec:additional_experiments}

In this section, we provide the expected coverage, nominal log posterior and balancing error for all the algorithm/benchmark pairs considered in this paper. We have added the Ratio Neural Posterior Estimation algorithm (RNPE) which corresponds to a normalizing flow trained as a ratio estimator obtained from the following transformation
\begin{equation}
    \varpi(y=1 \mid \btheta, \bx; \phat) \coloneqq \frac{\rhat(\btheta, \bx)}{1+\rhat(\btheta, \bx)} = \frac{\phat(\btheta, \bx) / p(\btheta)}{1+\phat(\btheta, \bx) / p(\btheta)}.
\end{equation}

\begin{figure}[h]
    \centering
    \includegraphics[width=\textwidth]{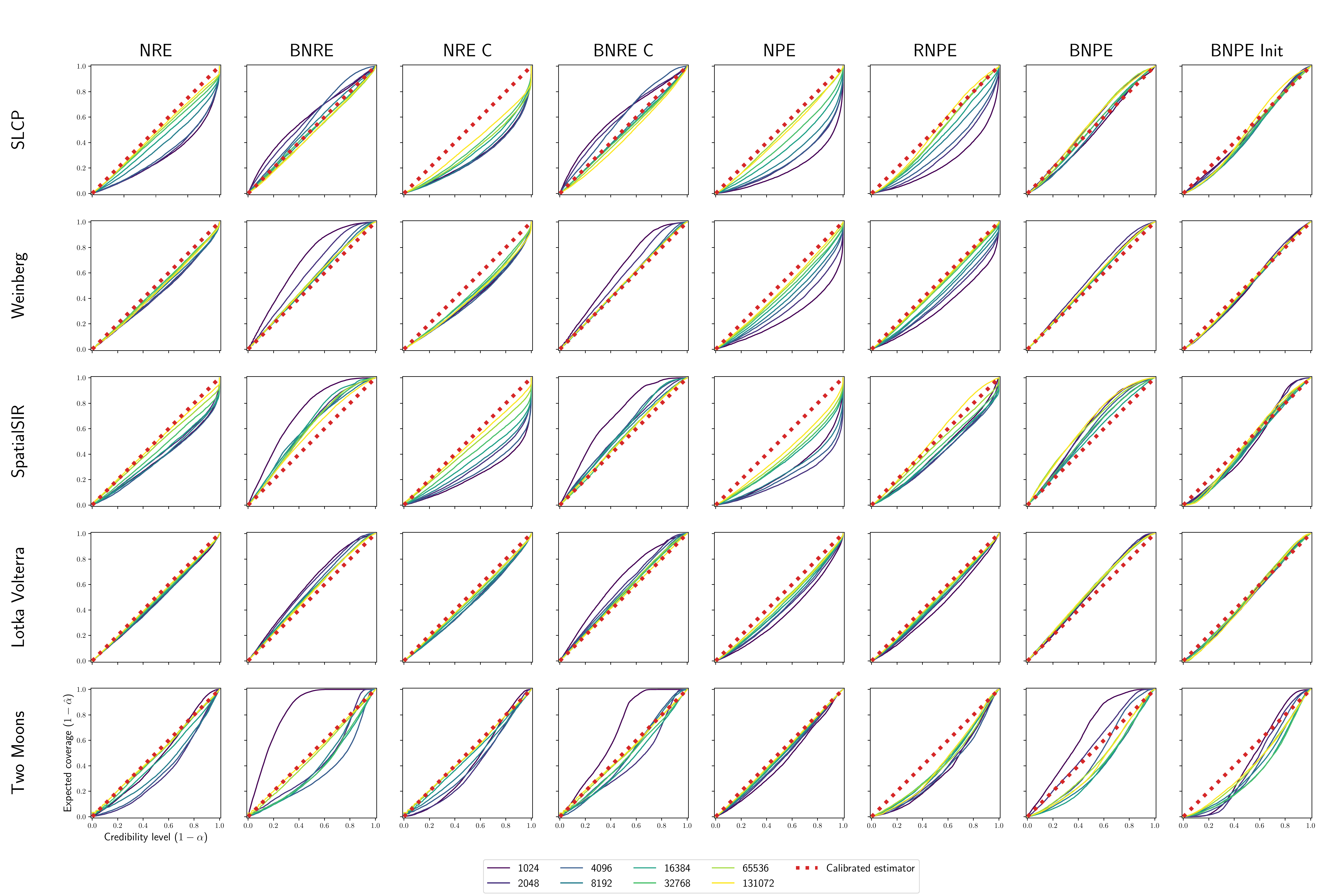}
    \caption{Expected coverage for increasing simulation budgets. A perfectly calibrated posterior has an expected coverage probability equal to the nominal coverage probability and hence produces a diagonal line. A conservative estimator has an expected coverage curve at or above the diagonal line, while an overconfident estimator produces curves below the diagonal line. 5 runs are performed for each simulation budget and the median is reported.}
    \label{fig:coverage_2}
\end{figure}

The expected coverage is defined as 
\begin{equation*}
    \E_{p(\btheta, \bx)} \left[ \indicator(\btheta \in \Theta_{\phat(\btheta \mid \bx)}(1 - \alpha) ) \right],
\end{equation*}
and is shown in \figureref{fig:coverage_2}.

\begin{figure}[h]
    \centering
    \includegraphics[width=\textwidth]{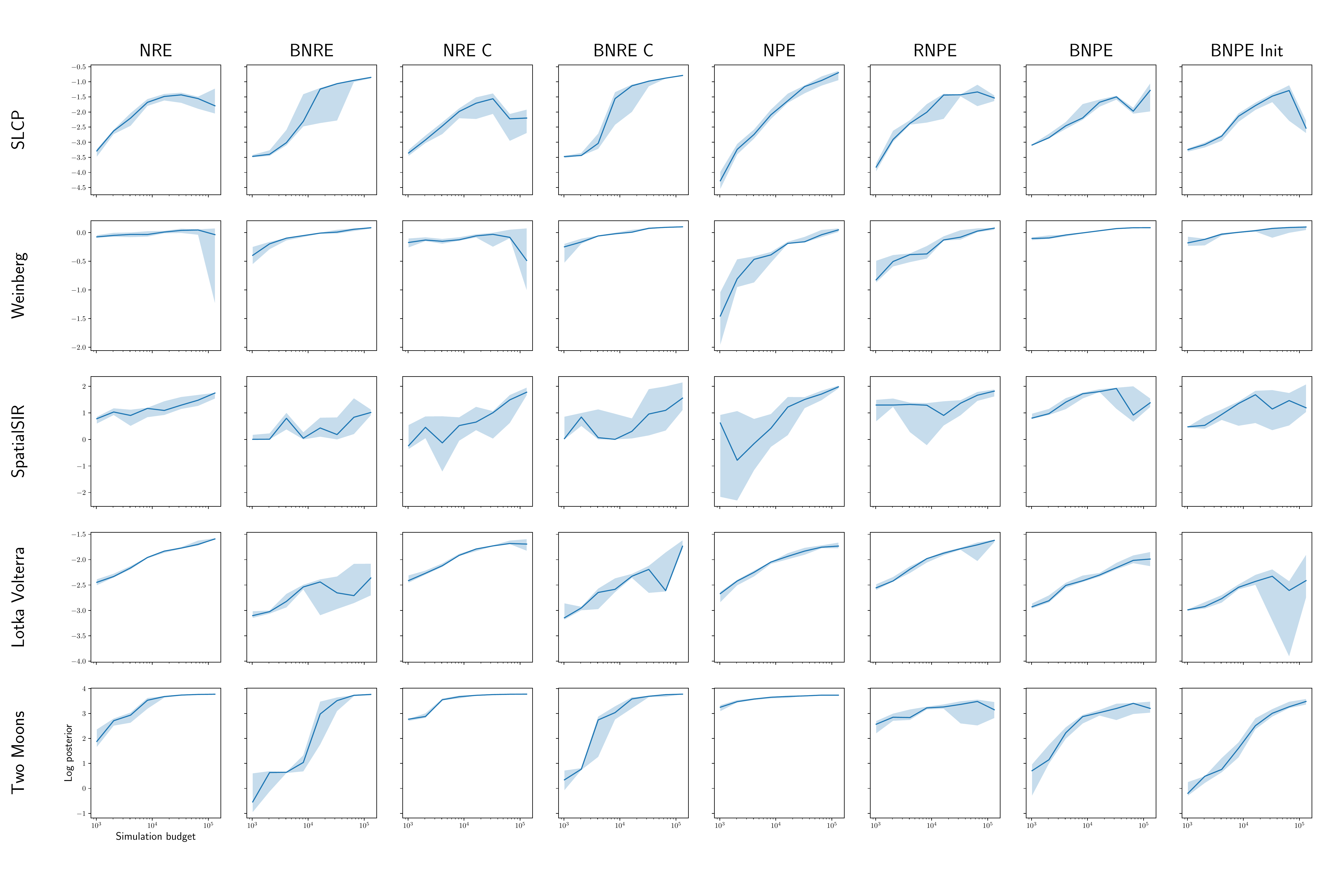}
    \caption{Nominal log posterior for increasing simulation budgets. 5 runs are performed for each simulation budget. Solid lines represent the median and the shaded areas represent the minimum and maximum. Larger values are desirable.}
    \label{fig:log_posterior}
\end{figure}

The nominal log posterior is defined as
\begin{equation*}
    \E_{p(\btheta, \bx)} \left[ \log \phat(\btheta \mid \bx) \right],
\end{equation*}
and is shown in \figureref{fig:log_posterior}. We observe that balanced algorithms have a lower nominal log posterior than their corresponding non-balanced algorithms. However, they have similar nominal log posterior as the simulation budget increases.

\begin{figure}[h]
    \centering
    \includegraphics[width=\textwidth]{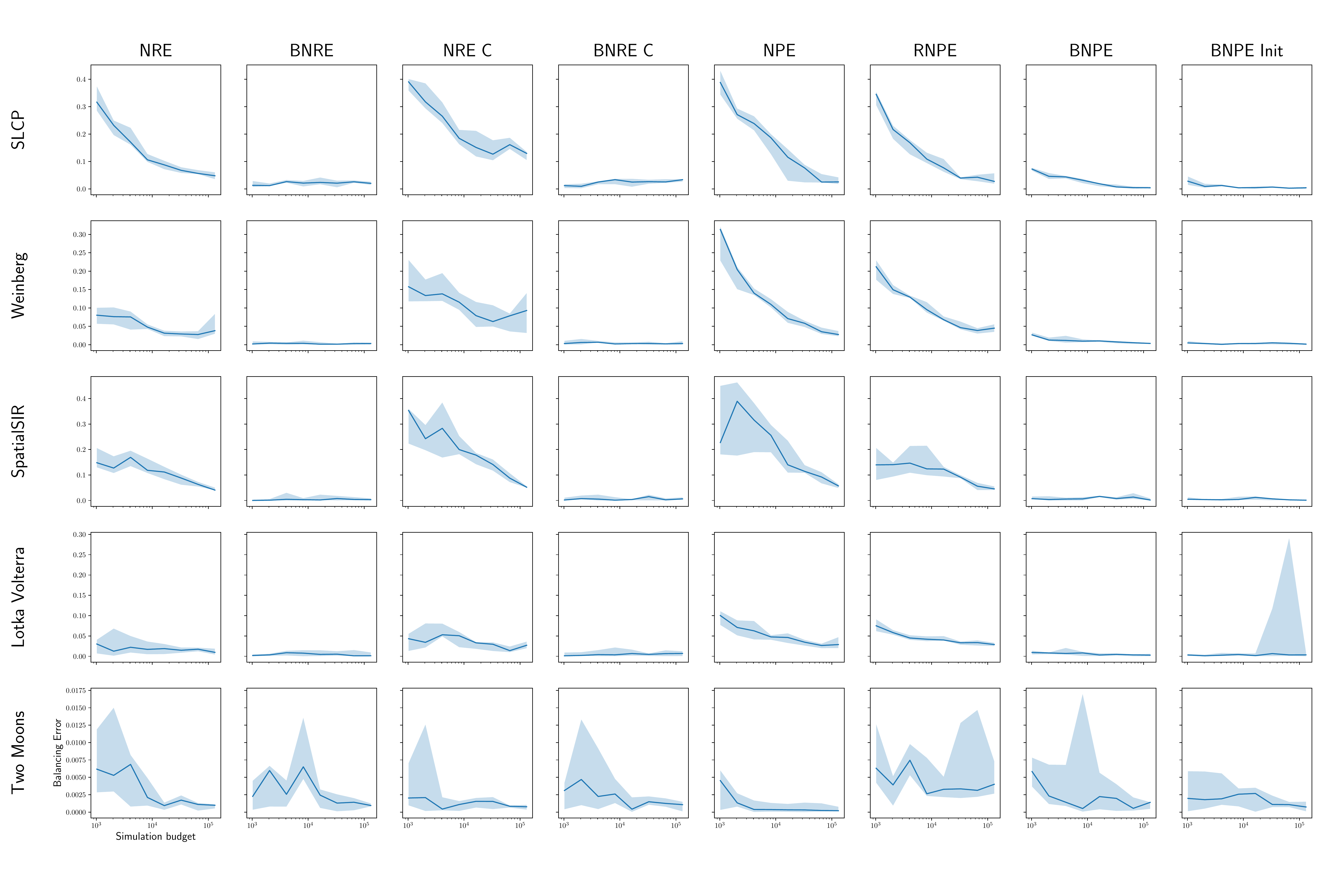}
    \caption{Balancing error for increasing simulation budgets. 5 runs are performed for each simulation budget. Solid lines represent the median and the shaded areas represent the minimum and maximum. Smaller values are desirable.}
    \label{fig:balancing_error_2}
\end{figure}

The balancing error is defined as
\begin{align*}
    & \left| \int (p(\btheta)p(\bx) + p(\btheta, \bx)) \varpi(y=1 \mid \btheta, \bx) \, d\btheta \, d\bx - 1 \right| \\
    =& \left| \int (p(\btheta)p(\bx) + p(\btheta, \bx)) \frac{\hat{p}(\btheta| \bx)/p(\btheta)}{1+\hat{p}(\btheta| \bx)/p(\btheta)} \, d\btheta \, d\bx - 1 \right|,
\end{align*}

and is shown in \figureref{fig:balancing_error_2}. Without surprise, we observe that enforcing the balancing condition leads to more balanced surrogates. Non-balanced algorithms show a high balancing error for low simulation budgets but this balancing error diminishes as the simulation budget gets higher. This is due to the fact that the posterior surrogate gets closer to the true posterior as the simulation increases and that the true posterior is always balanced.

\clearpage

\section{Qualitative posterior analysis}
\label{sec:qualitative}

\begin{figure}[h]
    \centering
    \includegraphics[width=0.9\textwidth]{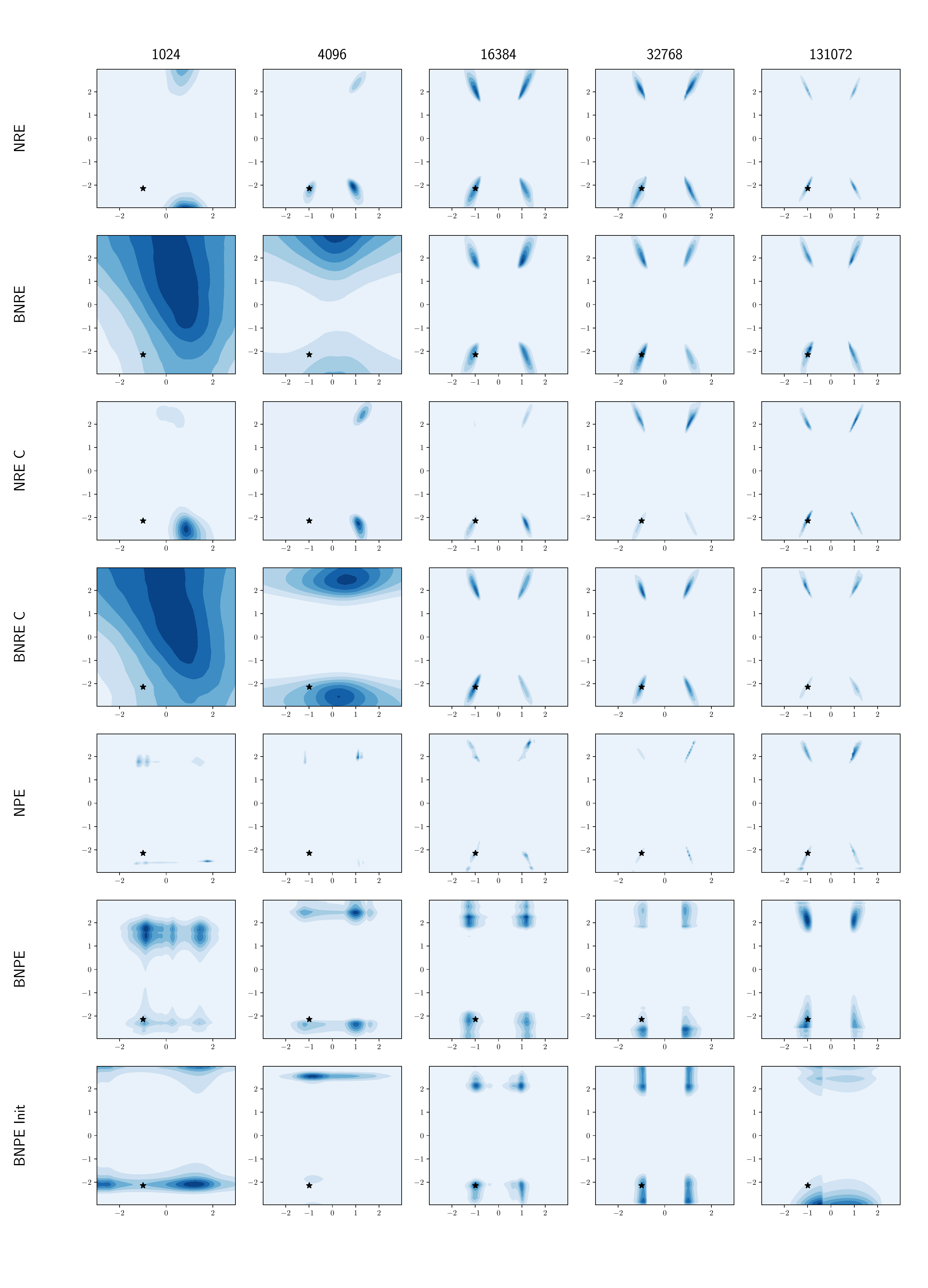}
    \caption{Visualization of posteriors obtained with different algorithms and simulation budgets on the SLCP benchmark. The blue areas represent the posterior and the black star represents the nominal parameters used for generating the observation.}
    \label{fig:slcp_posterior}
\end{figure}

\begin{figure}[h]
    \centering
    \includegraphics[width=0.9\textwidth]{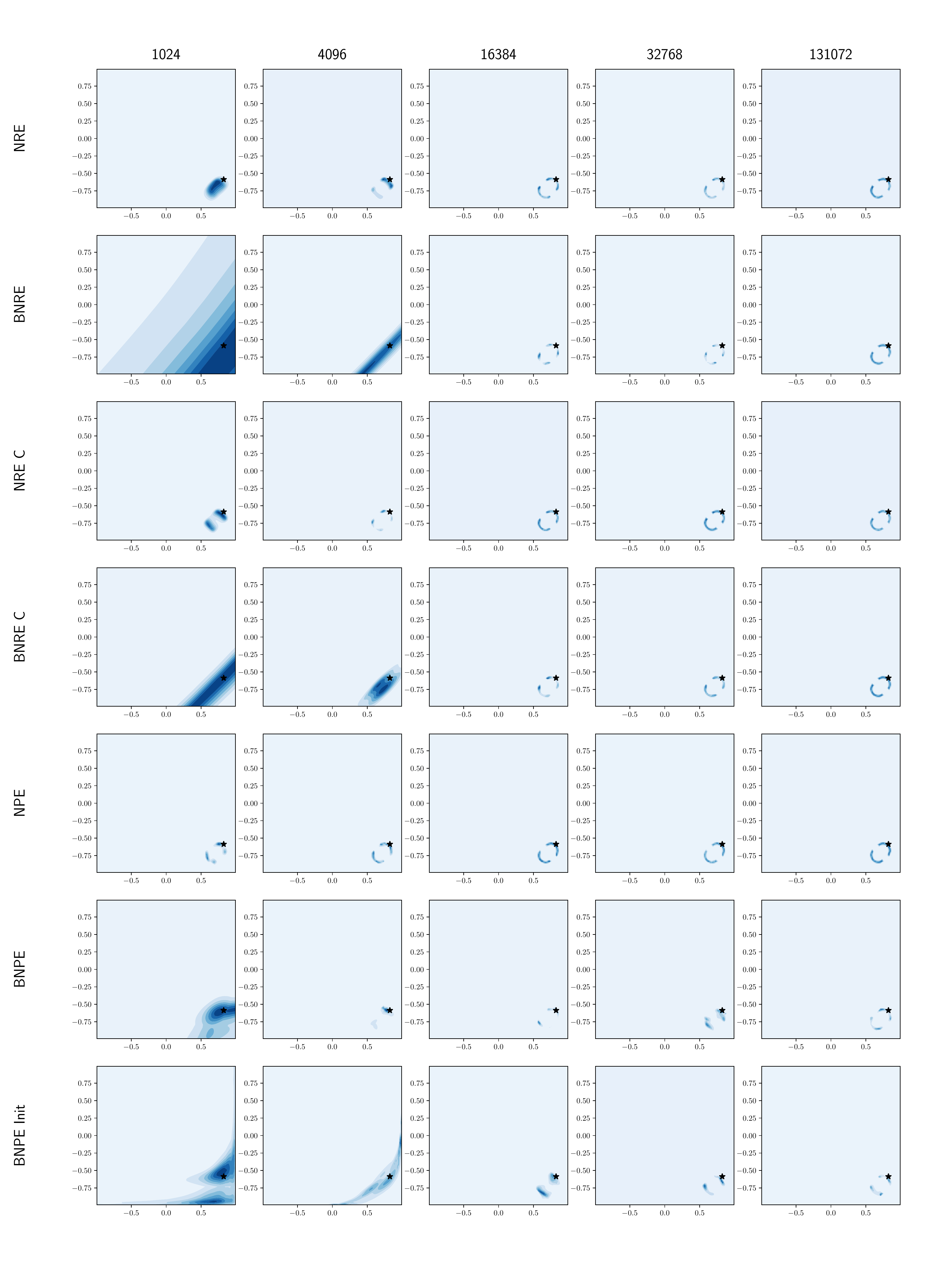}
    \caption{Visualization of posteriors obtained with different algorithms and simulation budgets on the two moons benchmark. The blue areas represent the posterior and the black star represents the nominal parameters used for generating the observation.}
    \label{fig:two_moons_posterior}
\end{figure}

In this section, we provide a visualization of the posteriors obtained with the different algorithms and simulation budgets. A visualization of the SLCP benchmark which is provided in \figureref{fig:slcp_posterior} and the two moons benchmark is shown in Figure \ref{fig:two_moons_posterior}. We observe that balanced versions of the algorithms indeed produce larger posterior approximations than non-balanced ones for low simulation budgets. Balanced versions then include the nominal parameter more often. Nevertheless, Balanced algorithms' approximate posteriors shrink as the simulation budget increases. BNRE and BNRE-C recover similar posterior surrogates as their non-balanced versions. However, BNPE leads to larger posteriors on the SLCP benchmark. Let us note that the architectures used for classifier-based and flow-based methods are different. Using a more flexible flow architecture would certainly lead to narrower posterior approximations as the simulation budget increases.

\clearpage

\section{Refining Balanced Neural Posterior Estimation}
\label{sec:refining}

\begin{figure}[h]
    \centering
    \includegraphics[width=\textwidth]{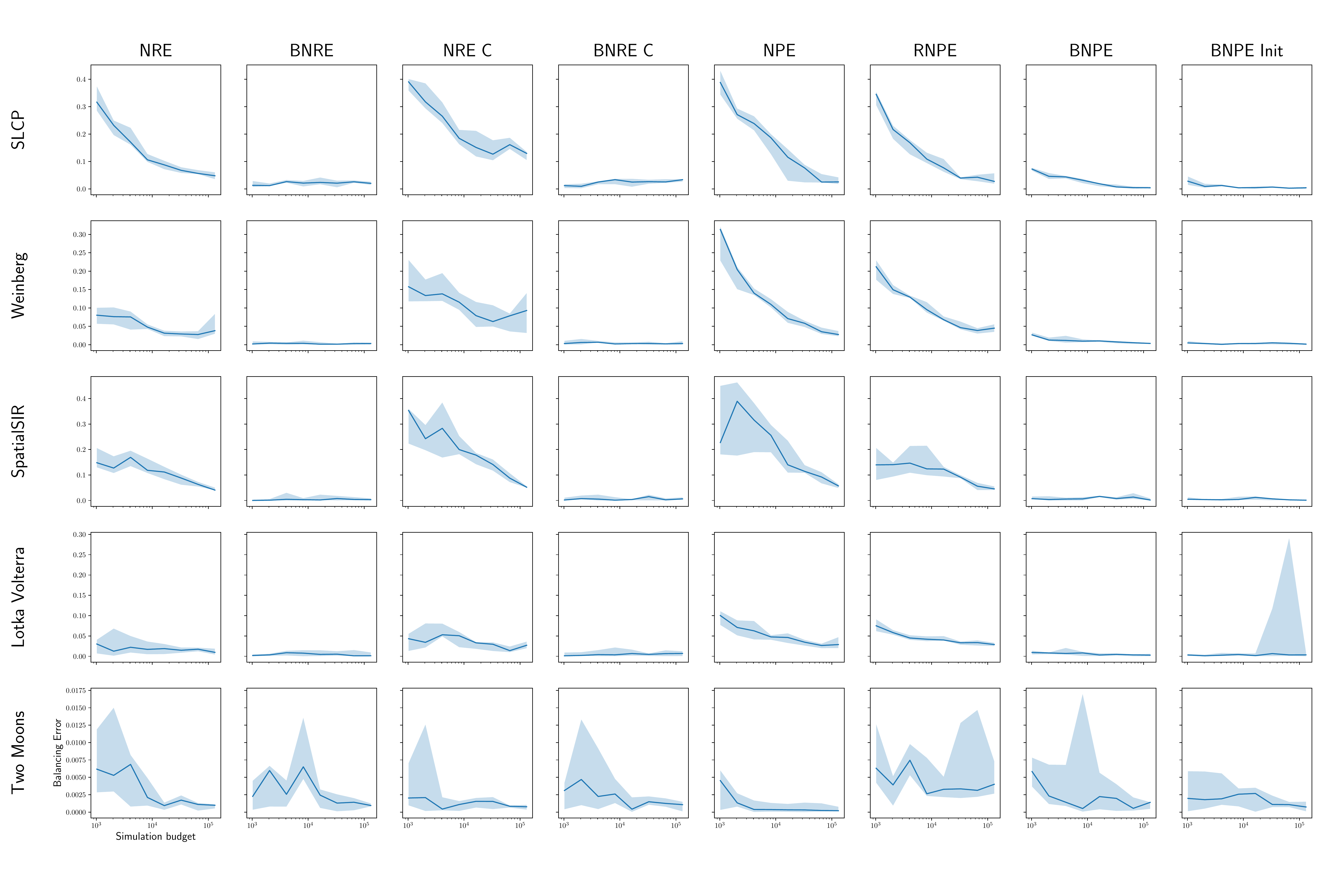}
    \caption{Comparison of different algorithms in terms balancing error $\left| \int (p(\btheta)p(\bx) + p(\btheta, \bx)) \varpi(y=1 \mid \btheta, \bx) \, d\btheta \, d\bx - 1 \right|$. 5 runs are performed for each simulation budget. Solid lines represent the median and the shaded areas represent the minimum and maximum. Lower values are desirable. Note: This is the same as Figure \ref{fig:balancing_error_2}.}
    \label{fig:balancing_error}
\end{figure}

\figureref{fig:balancing_error} shows the balancing error defined as
\begin{equation*}
    \left| \int (p(\btheta)p(\bx) + p(\btheta, \bx)) \varpi(y=1 \mid \btheta, \bx) \, d\btheta \, d\bx - 1 \right|.
\end{equation*}
We observe that while BNRE and BNRE-C show a lower balancing error on all the benchmarks, BNPE sometimes has a high balancing error. This suggests that BNPE sometimes struggles to learn to be balanced. In this section, we provide a way to simplify learning.

\subsection{Improving Balanced Neural Posterior Estimation}

Let us first observe that the prior $p(\btheta)$ is balanced:
\begin{align}
    & \int (\pi(\btheta, \bx \mid y=0) + \pi(\btheta, \bx \mid y=1)) \varpi(y=1 \mid \btheta, \bx) d\btheta \, d\bx \\
    =& \int (p(\btheta)p(\bx) + p(\btheta, \bx)) \frac{\hat{p}(\btheta| \bx)/p(\btheta)}{1+\hat{p}(\btheta| \bx)/p(\btheta)} d\btheta \, d\bx \\
    =& \int (p(\btheta)p(\bx) + p(\btheta, \bx)) \frac{p(\btheta)/p(\btheta)}{1+p(\btheta)/p(\btheta)} d\btheta \, d\bx = 1
\end{align}

BNRE can easily model the prior as this is achieved for $\varpi(y=1 \mid \btheta, \bx) = 0.5 \quad \forall \btheta, \bx$. In opposition, BNPE has to explicitly learn a balanced distribution. In order to ease the task of learning a balanced distribution, we propose to initialize the posterior surrogate to the prior distribution. In the case of a normalizing flow, this can be achieved by initializing all the transformations to the identity function and either using the prior as base distribution or adding a transform that maps the base distribution to the prior at the end of the flow. We will refer to this algorithm as Initialized Balanced Neural Posterior Estimation (BNPE Init). In this work, we use Neural Spline Flows \citep{durkan2019neural}. We discuss how to initialize this architecture to the prior in Appendix \ref{sec:init}.

It can be observed from Figure \ref{fig:balancing_error} that initializing the posterior surrogate to the prior indeed leads to a lower balancing error. Let us note that although the posterior surrogate is initialized to the prior, with a high enough simulation budget, it is able to learn a balanced surrogate that carries information about the parameter as seen from the log posterior quantity in Figure \ref{fig:log_posterior}. 

\subsection{Intializing neural spline flows to the prior distribution}
\label{sec:init}

A normalizing flow models a complex distribution as a sequence of transformations of some base distribution. Consequently, the flow models the prior in the two following scenarios:
\begin{itemize}
    \item the transformations are all identity transformations and the base distribution is the prior, or
    \item the transformations are all identity transformations and a fixed transformation that maps the base distribution to the prior is added at the end of the flow.
\end{itemize}
In the following, we describe how to obtain the two core components: identity transformations and a transformation that maps the base distribution to the prior. We then discuss the advantages and drawbacks of the two methods. In the experiments, we used a transformation from the base distribution to the prior.

\paragraph{Neural spline transformations} \citep{durkan2019neural} are transformations defined by $K$ rational-quadratic functions, with boundaries set by $K+1$ knots denoted by $(x^{(k)}, y^{k})_{k=0}^K$. The boundaries are defined by $(x^{(0)}, y^{0}) = (-B, -B)$ and $(x^{(K)}, y^{K}) = (B, B)$. The knots are parametrized by two vectors $\theta^w$ and $\theta^h$ of length $K$. Those vectors are passed through a softmax and multiplied by $2B$ to define the bins' width and height. 

The rational-quadratic in the $k^{\text{th}}$ bin is defined as 
\begin{equation}
    \frac{\alpha^{(k)}(\xi)}{\beta^{(k)}(\xi)} = y^{(k)} + \frac{(y^{(k+1)} - y^{(k)})[s^{(k)}\xi^2 + \delta^{(k)}\xi(1-\xi)]}{s^{(k)} + [\delta^{(k+1)} + \delta^{(k)} - 2s^{(k)}] \xi(1-\xi)}.
\end{equation}
The terms $\left\{\delta^{(k)}\right\}_{k=1}^{K-1}$ define the derivatives at the internal points and are parametrized by the vector $\theta^d$ of length $K-1$. The other terms are defined as $s_k = (y^{k+1} - y^k)/(x^{k+1} - x^k)$ and $\xi = (x - x^k)/(x^{k+1} - x^k)$.

\paragraph{Initializing transformations to identity}
To let the spline transformation be close to identity, we need to initialize the knots such that all the weights and widths are the same. This can be done by initializing the vectors $\theta^w$ and $\theta^h$ to zeros for all conditionings. In addition, all $s^{(k)}$ are then equal to $1$. We also need $\delta^{(k)} = 1, \ \forall k$. In the implementation used \citep{Rozet_Zuko_2022}, the vector $\theta^d$ models the log derivatives and hence must be initialized to a vector full of zeros.

To achieve the identity transform, the outputs of the neural network $\theta^h$, $\theta^w$ and $\theta^d$ must be zeros at initialization for all conditionings. This is achieved when biases and weights are all set $0$. However, this initialization is not optimal for learning via stochastic gradient descent, so we aim for a trade-off between ease of training and closeness to the prior. We initialize the neural network using some standard initialization, set the biases to $0$ and divide the weights by $5$ to obtain a function close to an identity transformation.

\paragraph{Mapping the base distribution to the prior}
In our case, the base distribution is a normal distribution $\mathcal{N}(0,1)$ and the priors are uniform in all the benchmarks considered. Therefore, we need to define a mapping from a Uniform distribution $\mathcal{U}(a, b)$, which p.d.f. is denoted by $p_u(u)$, to a normal distribution $\mathcal{N}(0, 1)$, which p.d.f. is denoted by $p_n(n)$ and c.d.f denoted $F_n(n)$. Let us first consider an intermediate mapping to a Uniform distribution $\mathcal{U}(0, 1)$ which p.d.f is denoted $p_{\tilde{u}}(\tilde{u})$. 
Going from $u$ to $\tilde{u}$ can be achieved with the following transformation
\begin{equation}
    \tilde{u} = \frac{u + a}{b - a}.
\end{equation}
The Jacobian linked to this transformation is then
\begin{equation}
    \frac{1}{b - a}
\end{equation}

Going from $\tilde{u}$ to $n$ can be achieved with the following transformation
\begin{equation}
    \tilde{u} = F_n(n) \Leftrightarrow n = F_n^{-1}(\tilde{u}).
\end{equation}
The Jacobian linked to this transformation is then
\begin{align}
    \left| (F_n^{-1})'(\tilde{u}) \right| &= \left| \frac{1}{F_n'(F_n^{-1}(\tilde{u}))} \right|\\
    &= \frac{1}{p_n(F_n^{-1}(\tilde{u}))}.
\end{align}

\paragraph{Comparison of the different initialization schemes} 
We have considered two initialization schemes: using the prior as base distribution or adding a transformation that maps the base distribution to the prior distribution. Using the prior distribution as base distribution has the advantage to be easy to implement as it does not require defining a mapping between both distributions. However, we have empirically observed that modifying the base distribution can lead to worse performance. We hypothesize that this is due to the fact that all the considered benchmarks have a uniform prior while flows work better with a Gaussian base. Let us note that using such a transformation could be beneficial in the more general setting as it solves leakage! Leakage is the fact that NPE algorithms can lead to a posterior surrogate that has density outside of the prior support. This is a problem in sequential settings where this surrogate is used as a proposal for simulating new data points. The transformation from a normal distribution to the prior is a transformation that maps any distribution with infinite support to a distribution that has the same support as the prior. Therefore, leakage cannot happen when such a transformation is applied.

\end{document}